\documentclass[runningheads]{llncs}
\usepackage{graphicx}
\usepackage{tikz}
\usepackage{comment}
\usepackage{amsmath,amssymb, bm} % define this before the line numbering.
\usepackage{color}

\usepackage{hyperref}
\hypersetup{colorlinks}
\usepackage{epsfig}
\usepackage{graphicx}
\usepackage{tabularx,booktabs}
\usepackage{multirow}
\usepackage{wrapfig}
\newcolumntype{C}{>{\centering\arraybackslash}X} % centered version of "X" type
\newcolumntype{S}{>{\small}c}
\usepackage{adjustbox}
\usepackage[accsupp]{axessibility}  % Improves PDF readability for those with disabilities.

\usepackage{overpic}
\usepackage{mdframed}

\usepackage[width=122mm,left=12mm,paperwidth=146mm,height=193mm,top=12mm,paperheight=217mm]{geometry}

\usepackage{xcolor}
\definecolor{light-gray}{gray}{0.95}

\newcommand{\etal}{\textit{et al.}}
\newcommand{\reals}{\mathbb{R}}
\newcommand{\bin}{\bm{b}}
\newcommand{\binij}{b_{ij}}
\newcommand{\bbox}{\mathcal{B}}
\newcommand{\query}{\mathcal{Q}}

\newcommand{\appx}{{\raise.17ex\hbox{$\scriptstyle\sim$}}}
\newcommand{\mdelta}[1]{\textbf{$\delta_#1$}$\uparrow$}

\usepackage{stackengine}
\usepackage{scalerel}
\usepackage{fp}
\def\clipeq{\!\mathrm{=}\!}
\def\Lla{\Longleftarrow\!\!\!\!\!\!}
\def\Lra{\!\!\!\!\!\!\Longrightarrow}
\newcount\argwidth
\newcount\clipeqwidth
\savestack\tempstack{\stackon{$\clipeq$}{}}%
\clipeqwidth=\wd\tempstackcontent\relax
\newcommand\xleftrightarrows[2]{%
  \savestack\tempstack{\stackon{$\scriptstyle#1$}{$\scriptstyle#2$}}%
  \argwidth=\wd\tempstackcontent\relax%
  \FPdiv\scalefactor{\the\argwidth}{\the\clipeqwidth}%
  \FPsub\scalefactor{\scalefactor}{1.7}% <---CAN PLAY WITH THIS VALUE
  \FPmax\scalefactor{\scalefactor}{.05}%
  \mathrel{%
  \stackunder[2pt]{\stackon[3pt]{$\Lla\hstretch{\scalefactor}{\clipeq}\Lra$}%
     {$\scriptstyle#1$}}{$\scriptstyle#2$}%
  }%
}

\usepackage{pifont}% http://ctan.org/pkg/pifont
\newcommand{\cmark}{\ding{51}}%
\newcommand{\xmark}{\ding{55}}%

\begin{document}
% \linenumbers
\pagestyle{headings}
\mainmatter
\def\ECCVSubNumber{****}  % Insert your submission number here

% Proposed title
\title{LocalBins: Improving Depth Estimation by Learning Local Distributions} % Replace with your title

\begin{comment}
\titlerunning{ECCV-22 submission ID \ECCVSubNumber} 
\authorrunning{ECCV-22 submission ID \ECCVSubNumber} 
\author{Anonymous ECCV submission}
\institute{Paper ID \ECCVSubNumber}

\end{comment}
%******************

% \begin{comment}
\titlerunning{LocalBins: Improving Depth Estimation by Learning Local Distributions}
%

\authorrunning{S. F. Bhat et al.}
\author{Shariq Farooq Bhat\inst{1} \and
Ibraheem Alhashim\inst{2} \and
Peter Wonka\inst{1}}

\institute{KAUST \and
National Center for Artificial Intelligence (NCAI), Saudi Data and Artificial Intelligence Authority (SDAIA), Riyadh, Kingdom of Saudi Arabia\\
\email{\tt \scriptsize shariq.bhat@kaust.edu.sa, ibraheem.alhashim@gmail.com, pwonka@gmail.com}
}
% \end{comment}

%******************

\maketitle

\begin{abstract}
We propose a novel architecture for depth estimation from a single image. The architecture itself is based on the popular encoder-decoder architecture that is frequently used as a starting point for all dense regression tasks.
We build on AdaBins which estimates a global distribution of depth values for the input image and evolve the architecture in two ways. 
First, instead of predicting global depth distributions, we predict depth distributions of local neighborhoods at every pixel.
Second, instead of predicting depth distributions only towards the end of the decoder, we involve all layers of the decoder.
We call this new architecture LocalBins. Our results demonstrate a clear improvement over the state-of-the-art in all metrics on the NYU-Depth V2  dataset. Code and pretrained models will be made publicly available.\footnote{\url{https://github.com/shariqfarooq123/LocalBins}}

\end{abstract}

\section{Introduction}
In this paper, we propose a new architecture for learning to estimate depth values given a single input image. In this line of work there are two main approaches that have been followed recently. Combining to train on multiple datasets at once while factoring out scale, e.g.~\cite{Gordon_2019_ICCV,Ranftl2020MiDaS} and training on a single dataset with consistent scale, e.g.~\cite{Eigen2014,Laina2016,Xu2017,Hao2018DetailPD,Xu2018StructuredAG,Fu2018DeepOR,Hu2018RevisitingSI,Alhashim2018,bts_lee2019big,dav_huynh2020guiding,Bhat2021}. Our approach falls in the second category. There are multiple published competing architectures, with AdaBins~\cite{Bhat2021} currently being the most successful architecture on datasets such as NYU-Depth V2~\cite{Silberman2012}. Our newly proposed architecture, called \emph{LocalBins} aims to improve upon this work.

The main idea of AdaBins is to predict adaptive bins that estimate one ``global" depth distribution per image. This prediction works both as auxiliary supervision of depth estimation, but also directly influences the depth prediction.

We initially formulated two objectives in evolving AdaBins. First, we wanted to see if predicting local distributions around each pixel can improve upon predicting one global distribution for the complete input image. 
Second, we wanted to design the architecture such that depth distribution supervision can be injected earlier in the network, preferably in a multi-scale fashion. AdaBins needs a special architecture design to work. Estimating global adaptive bins needs a transformer at ``high resolution". Estimation of bins is done close to the output layer and most of the work is delegated to a specialized module based on a transformer. Even though this improves performance significantly, this offload of work may prevent earlier layers to fully exploit the ``distribution supervision" to learn better representations. We call this the `late injection problem' in our arguments. Any attempts to estimate global adaptive bins earlier in the network (e.g. near the bottleneck) or without a transformer leads to unstable training - divergence or convergence to a suboptimal point.

\begin{figure}[t]
    \centering
    \begin{overpic}[width=\textwidth,percent]{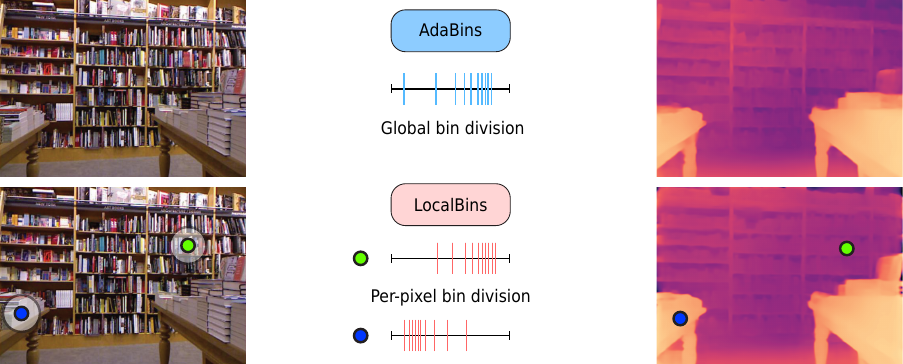}
    \end{overpic}
    \caption{Illustration of global adaptive bins vs local adaptive bins. While AdaBins predicts a depth distribution for a complete image, LocalBins predicts a depth distribution for the neighborhood of each pixel.}
    \label{fig:my_label}
\end{figure}

We realize these ideas in the following manner. To perform local predictions of depth distributions, we propose to use bin estimation at every pixel, and impose regularization on bin predictions via a query and response training scheme.
Our proposed module is regularized to predict the depth distributions within the randomly selected bounding boxes within the image.

To perform multi-scale predictions of depth distributions, we let the network predict local depth distributions in a gradual step-wise manner throughout the decoder. Starting with a small $N_{seed}$ number of bins, at the bottleneck each bin is subsequently split into two at every decoder layer, i.e., the $i^{th}$ layer of the decoder estimates $2^iN_{seed}$ bins at every pixel position. Together with locality, this coarse-to-fine construction lets us avoid unstable training and simultaneously solves the late injection problem. 

Our proposed LocalBins module is lightweight (adding only $\appx$1M params) and can be used in conjunction with any encoder-decoder network.

To summarize, we make the following contributions:
\begin{itemize}
    \item We propose a new architecture for single image depth estimation that improves upon the state-of-the-art in all metrics on the NYU-Depth V2~\cite{Silberman2012} dataset. Models and code will be made publicly available.
    \item We propose two novel architecture ideas to single image depth estimation: 1) estimating local histograms instead of a single global histogram and 2) estimating histograms in a multi-scale fashion to benefit from distribution supervision earlier in the pipeline. Even beyond depth estimation, we are not aware of existing similar concepts and we believe that these ideas could be beneficial beyond depth estimation.
\end{itemize}

\section{Related Work}
There are multiple categories of depth estimation methods. The first category are unsupervised methods~\cite{Xie2016Deep3DFA,Zhou2017,Godard2017,Godard2018DiggingIS,casser2019unsupervised,Gordon_2019_ICCV,li2020unsupervised,Zhou2021Diffnet,Watson_2021_CVPR}. These methods do not use ground truth depth data, but use self-supervision generally by some form of 3D reconstruction to learn depth values. These methods typically use videos or stereo videos as input.
The second category of methods learn depth estimation from multiple datasets~\cite{MDLi18,Ranftl2020MiDaS,Ranftl_2021_ICCV} jointly. Combining multiple datasets requires considering the different depth scales of the scenes. Therefore, methods that train on multiple datasets are generally not comparable to methods that train on a single dataset, because the test protocol is different. The third category of algorithms are domain transfer methods~\cite{atapour2018real,zhao2019geometry,Tonioni2019,CrDoCo2019,AkadaBAW22}. These techniques assume the availability of ground truth data in one domain during training, but the images in the target domain do not have ground truth depth (or only a few of them have ground truth depth~\cite{Zhao_2020_CVPR}). 
The fourth category are depth estimation methods that learn a depth estimation network for each dataset separately \cite{Eigen2014,Laina2016,Xu2017,Hao2018DetailPD,Xu2018StructuredAG,Fu2018DeepOR,Hu2018RevisitingSI,Alhashim2018,bts_lee2019big,dav_huynh2020guiding,Bhat2021}. Our method belongs to this category of supervised monocular depth estimation. These methods formulate the task as the regression of a depth map from a single RGB input image. The current dominant architecture follows an encoder-decoder network. Most high performing methods apply such architecture with some variations on the process of extracting of relevant feature maps during encoding and the fusion of these features with the intermediate maps produced during the decoding stage. 
Finally, we mention two very recent arXiv submissions that are concurrent to our work for the sake of completeness. The first is GLP-depth~\cite{kim2022global} which proposes a hierarchical transformer encoder that captures global features and a simple decoder that considers the local context. The second method \cite{Yuan2022_cvpr_crf} employs a neural window fully-connected Conditional Random Fields (CRFs) module for the decoder and a vision transformer for the encoder.
% %
% Unlike these concurrent works, our method does not require any vision transformer modules.
% In addition, there are techniques that try to use existing depth estimators to 

\section{Methodology}
\subsection{Background}
AdaBins~\cite{Bhat2021} divides the depth interval $(d_{min}, d_{max})$ into bins. This bin-division is global (one proposed bin-division per image) and adaptive (varies from image to image), and reflects the global depth distribution for the input image. Each bin can have a different size and the predicted bin centers are closer to each other near more frequently occurring depth values. AdaBins employs an encoder-decoder architecture followed by a transformer based module to predict the adaptive bins. The final depth estimation is obtained by predicting the pixel-wise probability distribution over the bins and computing an expectation over the predicted global bin centers. This can also be seen as expressing the final depth value as a linear combination of bin centers. The reader is referred to \cite{Bhat2021} for more details.

We build upon the basic idea of AdaBins to estimate depth distributions, but we change the architecture design to incorporate two novel ideas. First, instead of predicting a single global depth bin-division, our architecture estimates a bin-division at every pixel, reflecting the depth distribution in the local neighborhood. Thus, the bin-divisions not only can vary from image to image (adaptiveness) but also from pixel to pixel (locality). Second, we do not use a transformer as a subsequent separate architecture block but integrate the depth prediction more tightly in the decoder. We utilize all the layers of the convolutional decoder to gradually refine the bin-division proposed for each pixel. The details of our architecture are described in the next section.

\subsection{Architecture}
\label{subsec:arch}
\begin{figure}[t]
    % \begin{mdframed}
        \centering
        \begin{overpic}[width=\linewidth]{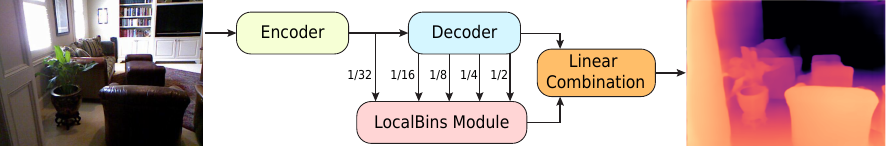}
        \end{overpic}
    % \end{mdframed}
    \caption{Architecture overview. LocalBins module contains pixel-wise operations and estimate local neighborhood bin density for each pixel location.
    }
    \label{fig:arch_overview}
\end{figure}

Our architecture has two major components (see Fig.~\ref{fig:arch_overview}): 1) a standard encoder-decoder block and 2) our proposed LocalBins module.

\textbf{Encoder-decoder} We use the same encoder-decoder architecture as AdaBins to facilitate a fair comparison (EfficientNet-B5 with skip connections).

\textbf{LocalBins module} The LocalBins module uses the bottleneck features and the decoder features from the encoder-decoder block to estimate the local distribution of depth values at every pixel. As in AdaBins, the estimated distributions are encoded as the adaptive bin-divisions of the depth range interval, with the density of resulting bin-centers directly reflecting the density of the depth values in the local neighborhood. In practice, the bin-divisions are formulated as a vector of normalized bin-widths at every pixel from which the bin-centers can be easily obtained via Eq.~\ref{eq:center_calc}. To estimate the bin-divisions at every pixel, we employ a \textit{coarse-to-fine binning} strategy. Starting with $N_{seed}$ number of bins for every pixel at the bottleneck, the number of bins doubles at every decoder layer.

\begin{figure}[t]
    \centering
    \begin{overpic}[width=\textwidth,percent]{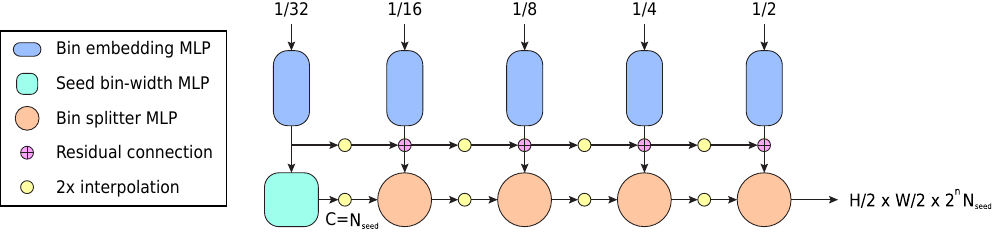}
    \end{overpic}
    \caption{Design of the LocalBins module. See Sec~\ref{subsec:arch} for more details.}
    \label{fig:localbins_module}
\end{figure}

The LocalBins module consists of three main types of layers: a) Bin embedding layers b) Seed bin width estimators c) Bin splitters. All these layers are composed of pointwise MLPs (a.k.a $1\times1$ convolutional blocks) with two hidden layers with hidden dimension $h$.
% (the choice of using only $1\times1$ ops will become clear in `Query-Response training'  section). 

\textbf{a) Bin embedding layers } All the feature blocks input to the LocalBins module (the bottleneck features from the encoder and the decoder features of all scales) are first fed to bin embedding layers. These layers project the features of varying channel dimensionality into the same space (dim=128) that we refer to as the `bin embedding space'. The further layers in the LocalBins module `decide' the bin-division for each pixel based on their bin-embeddings.

\textbf{b) Seed bin width estimator } This layer takes bin embeddings from the bottleneck as input and predicts $N_{seed}$ number of bins at each pixel of the bottleneck. This bin-division estimate is taken as the seed and subsequently each bin is divided into two bins for every subsequent decoder layer by the bin splitters (along with the spatial 2x interpolation).

\textbf{c) Bin splitters} These pointwise MLPs are used to realize our coarse-to-fine binning strategy. Loosely speaking, bin splitters `decide' where to put more bin-centers for each pixel based on their bin-embeddings. As illustrated in the Fig.~\ref{fig:localbins_module}, bin-embeddings and bin-widths from the previous layer are first bilinearly upsampled to match the spatial resolution of the current layer. A bin splitter MLP at layer $k$, denoted as $\mathcal{S}^k$, takes as input the `current' layer bin-embeddings after a residual connection with the upsampled previous layer bin-embeddings. The output is then used to split each bin-width from the previous layer into two. Specifically, let $\bin_{ij} \in \reals^{m}$ be the normalized $m$-bin-widths at pixel location $(i,j)$ (after 2x interpolation). Then, the new $2m$-bin-widths $\bin'_{ij} \in \reals^{2m}$ are given by:
\begin{equation}
    \bm{\alpha}_{ij} = \sigma(\mathcal{S}^k(\mathbf{e}_{ij}^{k-1} + \mathbf{e}_{ij}^k))
\end{equation}
\begin{equation}
    \bin'_{ij} = \{\alpha_{ij}^0 \binij^0, (1-\alpha_{ij}^0) \binij^0,\, \alpha_{ij}^1 \binij^1, (1-\alpha_{ij}^1) \binij^1,\,\dots,\alpha_{ij}^m \binij^m, (1-\alpha_{ij}^m) \binij^m\}
\end{equation}

where, $\sigma (\cdot)$ represents the splitter activation function that outputs values $\alpha \in (0,1)$, $\mathbf{e}_{ij}^k$ is the bin-embedding of the pixel at $(i,j)$ at the $k^{th}$ layer and $v^a$ represent the components of a vector $\mathbf{v}$.

We explore three designs of the splitter activation function $\sigma(\cdot)$, namely:
\begin{enumerate}
    \item Constant splitter: $\sigma(x)=0.5\,\, \forall x$, that divides a bin in half irrespective of the splitter MLP output (and indirectly the bin-embeddings). 
    \item Sigmoid splitter: where $\sigma(\cdot)$ represents the sigmoid activation function.
    \item Linear norm splitter: In this case, we let the splitter MLP $\mathcal{S}^k$ output two positive values $(x_1, x_2)$ (via ReLU) for each bin. Then the linear norm split is given by:
    \begin{equation}
        \sigma(x_1, x_2) = \frac{x_1}{x_1+x_2+\epsilon}
    \end{equation}
    where $\epsilon = 1e^{-4}$ is used for numerical stability. 
\end{enumerate}
Refer to Sec.~\ref{subsec:analysis} for their comparison.

Since the number of bins doubles every layer, we have $2^nN_{seed}$ bins at the end of an n-layer decoder. Therefore we use $output\_channels=2^nN_{seed}$ for the last convolutional layer in the decoder. We then use the hybrid regression as in AdaBins, to obtain the final depth map. The difference is that now the bins change from pixel-to-pixel:

\begin{equation}
    c(\binij^k) =  d_{min} + (d_{max} - d_{min})(\binij^k/2 + \sum_{s=1}^{k-1} \binij^s)
\end{equation}
\begin{equation}
\label{eq:center_calc}
    \Tilde{d}_{ij} = \sum_{k=1}^N c(\binij^k) p_{ij}^{k}
\end{equation}
where $\Tilde{d}_{ij}$ is the final estimated depth value, $\bin_{ij}$ and $\bm{p}_{ij}$ are the bin-widths at the output layer and softmax scores respectively at location $(i,j)$.

\label{sec:training}
\begin{figure}[t]
    \centering
    \includegraphics[width=0.99\linewidth]{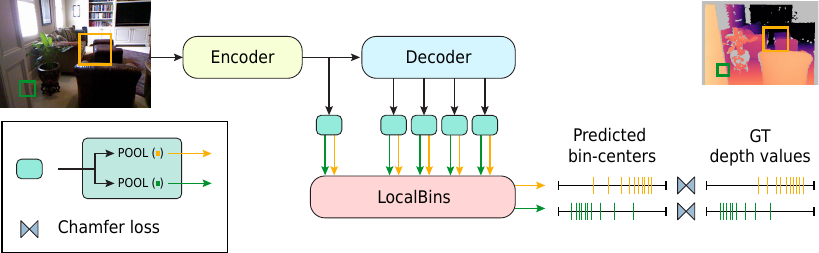}
    \caption{Illustration of Query-Response training. See Sec~\ref{subsec:qr_training} for more details.}
    \label{fig:qr_training}
\end{figure}

\subsection{Training}

Our network needs supervision in two forms. First, we need a pixel-wise loss ($\mathcal{L}_{pixel}$) to provide supervision for the final estimated depth values. For this we use the Scale-Invariant Loss as used in recent works~\cite{Bhat2021,bts_lee2019big}. Second, we need to supervise our network such that the bin predictions at a pixel actually reflect the density of depth values in its local neighborhood. 

A naive way would be to just choose a fixed-sized local window around every pixel (say a $5\times5$ window) and directly impose the regularization on bin-predictions at a pixel location such that they reflect the corresponding ground truth depth distributions within the window. However, there are two major problems with this approach: 1) It is not clear how one would choose the size of the window. Empirical determination is not a scalable solution as the amount of detail within a fixed sized window varies with the spatial resolution of the image. Alternatively, choosing different sizes simultaneously would lead to inconsistent regularization (e.g. bin predictions at the same pixel location would be compared to GT distributions within, say, $5\times5$ and $7\times7$ windows at the same time) 2) Chamfer loss, which is used in AdaBins to implement the distribution loss, is not computationally scalable. In AdaBins, it is computationally feasible because there is only one bin-division proposal per image. While in our case, we have a bin-division proposal at every pixel. This would mean we would have to compute Chamfer loss between the point sets at $\appx$300K locations per image (= total number of pixels) at the highest resolution, which is not feasible in terms of memory or computation. One can, in practice, subsample the number of locations. For example, for a batch size of 16, we could fit around $\appx$2\% of randomly selected pixel locations on four NVIDIA A100 GPUs. However, as expected, this leads to inferior performance (Sec.~\ref{subsec:analysis}). 

One obvious reason is the significant loss of the spatial coverage of locations at which the loss is computed. We initially identified two possible workarounds for this problem. Either investigate efficient ways for subsampling or let the gradients from the loss computation at a given pixel location directly flow to neighboring regions to increase the coverage. This work focuses on designing the latter solution. To increase the coverage, we propose to involve all the pixel locations within the window to compute the loss (instead of just the center one), while keeping the loss computation feasible. This means that instead of regularizing the bin-predictions at individual pixel locations, we propose to regularize the bin-predictions of the entire local window together, potentially solving the coverage problem. We achieve this via the following formulation that we call `Query-Response' training.

\subsubsection{Query-Response training}
\label{subsec:qr_training}
% Reader may have noted that we don't restrict or define the `area' of the `local neighborhood' in the design of architecture. Rather, 
We train the network via the following locality constraint regularization:

\textit{Consider a bounding box $\mathcal{B}$ at any given location in the input image. The LocalBins module, when applied on the spatial average of the features within $\mathcal{B}$, must predict the density of depth values within $\mathcal{B}$.}

This is illustrated in Fig.~\ref{fig:qr_training}. By using bounding boxes of different sizes, we can enforce the features to contain the local distributional information at multiple scales. Furthermore, since the spatial averaging operation covers the entire window, we can potentially achieve the complete coverage with a relatively smaller number of bounding boxes per-image.

In order to implement such a regularization we make use of ROIAlign aggregation~\cite{He2017_MaskRCNN}, a popular operation used in object detection pipelines. ROIAlign aggregation allows us to extract and pool the features at different layers (bottleneck and decoder features) corresponding to a given bounding box. Details on how ROIAlign works is found in \cite{He2017_MaskRCNN}. 

Given a bounding box $\bbox$ (\textit{a.k.a} query), we apply ROIAlign aggregation and use the pointwise MLPs (Bin embbeding, seed bin width estimator and bin splitter layers from the LocalBins module) on the pooled features to get the bins $\bin(\bbox)$ (\textit{a.k.a} the response).
% We denote this query-response operation by the shorthand notation $\bbox\xrightarrow[]{LBM}\bin$.
The resulting bins $\bin(\bbox)$ are then `forced' to match the ground truth depth distribution within that bounding box. We use the 1D bi-directional Chamfer loss as in ~\cite{Bhat2021} as the matching loss: $chamfer(\bin(\bbox),Depth(\bbox))$.

\subsubsection{``Foveated'' Loss}
We now have a few choices on how to aggregate losses from different bounding boxes to calculate the final loss. We choose to have smaller bounding boxes to have more influence than the larger ones. We therefore use the loss weights that exponentially decay with the bounding box size. Suppose we generate $N$ different sets of bounding boxes $\query_1, \dots, \query_N$ with different box sizes such that $\text{size}(\bbox_a\in \query_i)<\text{size}(\bbox_b\in \query_j)~\forall a,b \,\,\text{and} ~i~<~j$, the total loss is given by:
\begin{equation}
\label{eq:chamfer}
    \mathcal{L}_{bins} = \sum_{L=1}^{n}\gamma_l^{n-L}{\sum_{k=1}^{N}{\gamma_b^{k-1} \sum_{\bbox \in \query_k}{chamfer(\bin_L(\bbox), Depth(\bbox))}}}
\end{equation}

where $n$ is the number of layers and $b_L(\bbox)$ is the response at layer $L$ (running from bottleneck to output layer). We use $\gamma_l=\gamma_b = 0.3$ in our experiments and use 5 different sizes of bounding boxes as described below. In summary, we follow the steps: 
\begin{enumerate}
    \item Generate five sets of random bounding boxes of sizes, $3\times3$, $7\times7$, $15\times15$, $31\times31$, $63\times63$, each containing $M=200$ boxes.
    \item Extract the depth values from the ground truth depth map corresponding to these bboxes to use in the Chamfer loss calculation.
    \item Use ROIAlign with average pooling to get the corresponding pooled features at the bottleneck and decoder layers.
    \item Use bin embedding MLPs to get the corresponding bin-embeddings. At this stage, we have one bin-embedding vector at each layer corresponding to each bbox.
    \item Use seed bin estimator and bin splitter MLPs as usual to get the resulting bins.
    \item Calculate the unweighted Chamfer loss for the predicted bins against their ground truth (step 2).
    \item Calculate the weights and compute the weighted sum to get the final Chamfer loss, with the weighting scheme in Eq.~\ref{eq:chamfer}.
\end{enumerate}

Finally, we define the total loss as:
\begin{equation}
    \mathcal{L}_{total} = \mathcal{L}_{pixel} + \beta\mathcal{L}_{bins},
\end{equation}
where we set $\beta = 0.02$ in all our experiments.

\section{Implementation Details}
We implement the model in PyTorch~\cite{NEURIPS2019_bdbca288}. We use hidden dimension $h=256$ for the Seed bin width MLP and $h=128$ for Bin embedding MLPs and Bin splitter MLPs. We train the network with batch size of $16$ and use the AdamW~\cite{LoshchilovH19} optimizer with weight decay of $10^{-1}$. We use a learning rate of $3.57\times10^{-4}$ which is decayed by a factor of $10^{4}$ in the last $30\%$ iterations using a cosine decay schedule. We train the models for $10$ epochs in all our experiments.

\section{Experiments and Results}

\begin{figure}[t]
    \centering
    \begin{overpic}[width=\linewidth]{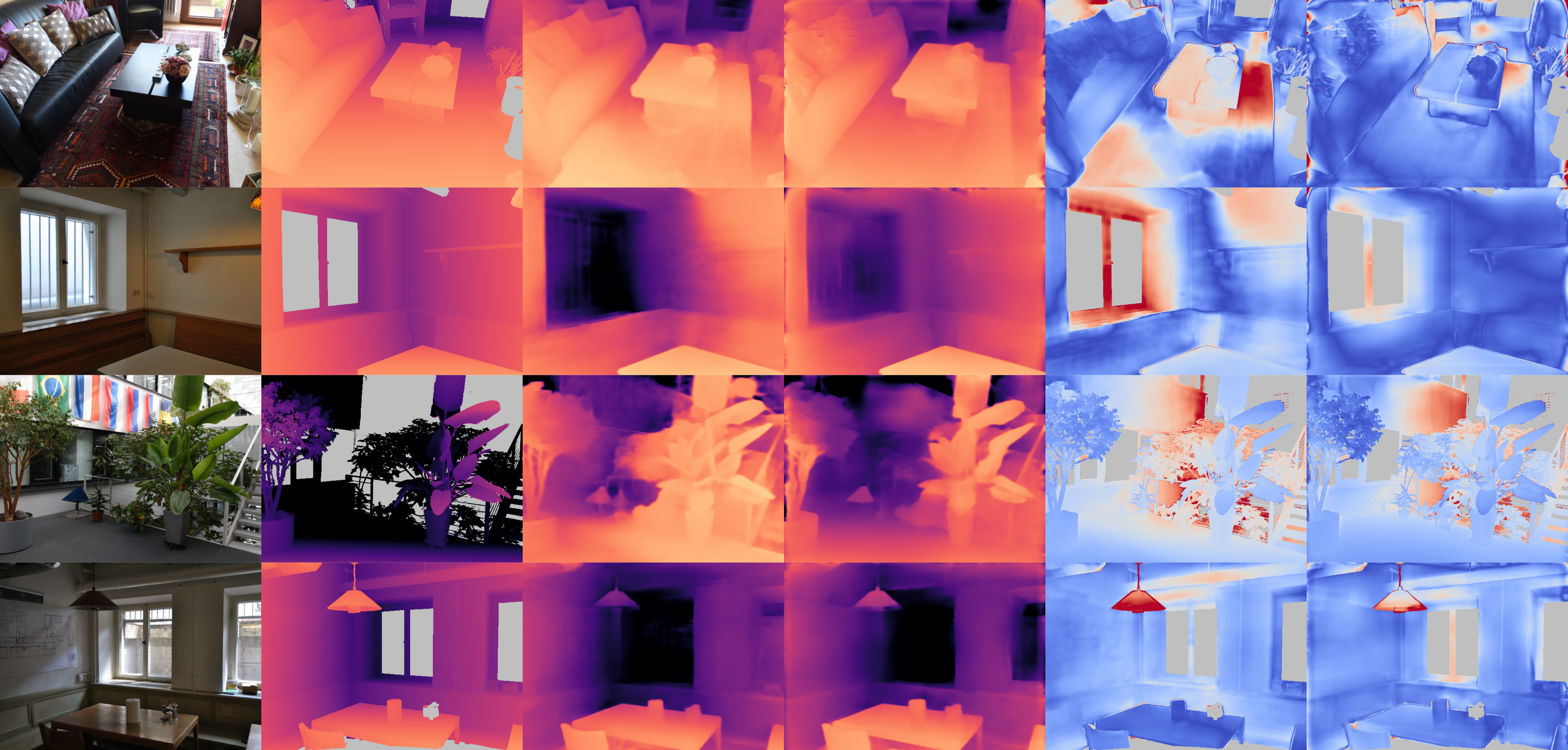}
    \put(6, -3){\small RGB}
    \put(23, -3){GT}
    \put(37, -3){AdaBins}
    \put(55, -3){Ours}
    \put(68.5, -3){AdaBins $\Delta$}
    \put(87, -3){Ours $\Delta$}
    \end{overpic}
    \caption{Qualitative results on iBims-1 benchmark without fine-tuning.}
    \label{fig:qualt_ibims}
\end{figure}

% We use the evaluation metrics as defined in previous works~\cite{}. \wip{TODO: NYU-Depth V2 and evaluation split and cropping strategy}
\subsection{Comparison to state-of-the-art}
\begin{figure}[t]
    \centering
    \begin{overpic}[width=\linewidth,tics=5]{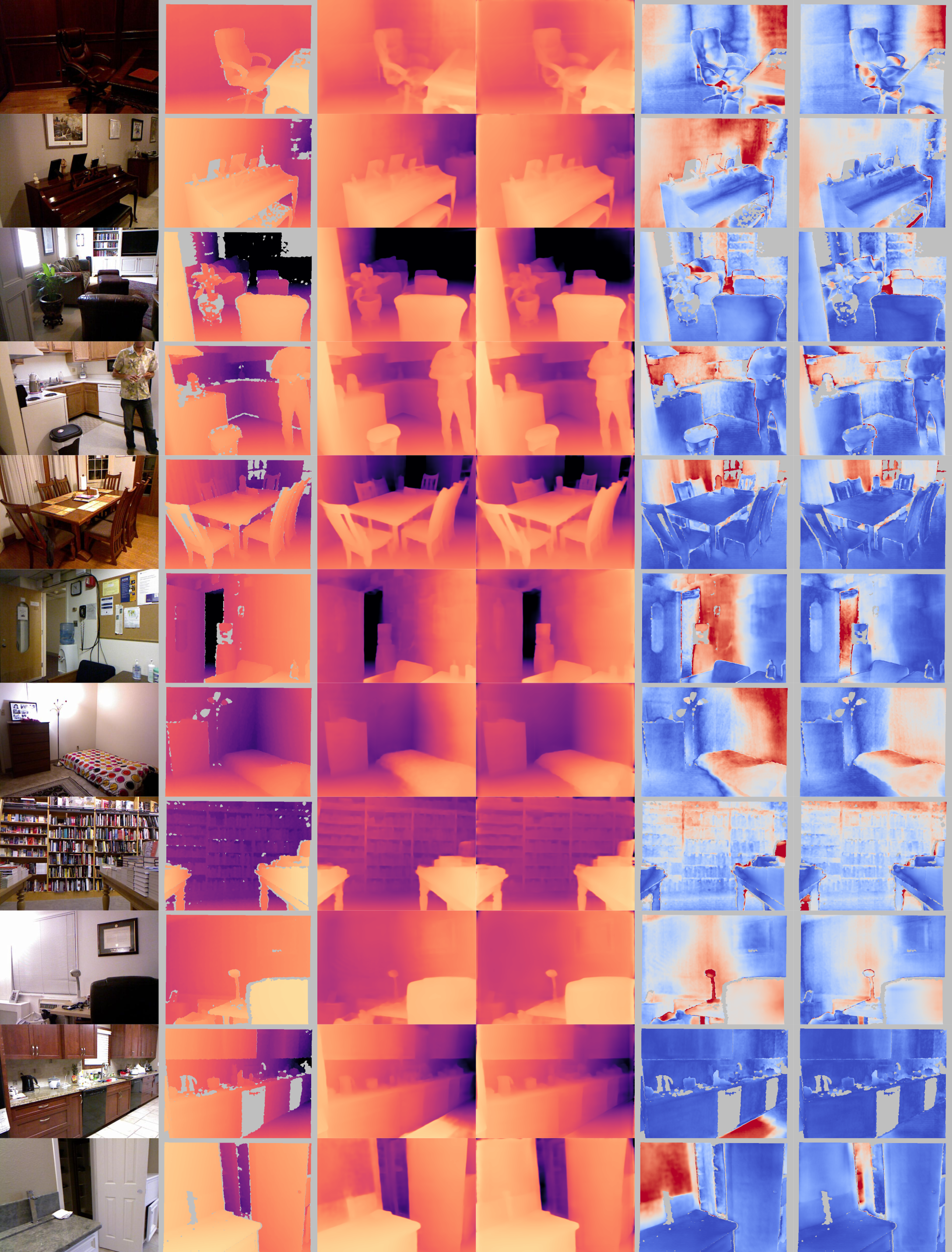}
    \put(4,102){RGB}
    \put(18,102){GT}
    \put(26,102){AdaBins~\cite{Bhat2021}}
    \put(41.5,102){Ours}
    \put(51,102){AdaBins $\Delta$}
    \put(65,102){Ours $\Delta$}
    
    \end{overpic}
    \caption{Qualitative results on NYU-Depth V2.}
    \label{fig:qualt-nyu}
\end{figure}

\textbf{NYU-Depth V2}. We use NYU-Depth V2~\cite{Silberman2012} as the most important dataset for evaluation. Table.~\ref{tab:results-nyu} presents the performance comparison on the official test set of NYU-Depth V2. Our proposed model shows the state-of-the-art performance across all metrics, with \appx4\% reduction in the absolute relative error metric. Qualitatively, as shown in Fig.~\ref{fig:qualt-nyu}, our model is better at predicting depths of thin objects as well as planar surfaces. Note that our final model is able to beat the current published state-of-the-art model AdaBins while having the same encoder-decoder backbone. In total, our model has even fewer parameters since we do not use global attention or transformers. We attribute this performance improvement to the proposed LocalBins design and better utilization of the depth statistics via our novel training scheme.

\begin{table}[t]
\centering
\begin{adjustbox}{width=\linewidth,center}
\begin{tabularx}{\linewidth}{@{}l|lS|*{5}{C}c@{}}
\toprule
\textbf{Method} & \textbf{Encoder}    & \textbf{\#\textit{p}}   & \textbf{$\delta_1$}$\uparrow$       & \textbf{$\delta_2$}$\uparrow$          & \textbf{$\delta_3$}~$\uparrow$            & REL~$\downarrow$          & RMS~$\downarrow$  & $log_{10}$~$\downarrow$ \\ \midrule
Eigen~\etal~\cite{Eigen2014} & - & 141                                                      & 0.769          & 0.950          & 0.988          & 0.158            & 0.641          & --              \\ 
Laina~\etal~\cite{Laina2016} & ResNet-50  & 64                                                  & 0.811          & 0.953          & 0.988          & 0.127            & 0.573          & 0.055                \\ 
% MS-CRF~\cite{Xu2017}                                                             & 0.811          & 0.954          & 0.987          & 0.121            & 0.586          & 0.052          \\ 
Hao~\etal~\cite{Hao2018DetailPD} & ResNet-101  & 60                                                     & 0.841          & 0.966          & 0.991          & 0.127            & 0.555          & 0.053                \\ 
Lee~\etal~\cite{Lee2011} & - & 119                                                      & 0.837          & 0.971          & 0.994          & 0.131            & 0.538          &   --                   \\
Fu~\etal~\cite{Fu2018DeepOR} & ResNet-101 & 110                                                     & 0.828          & 0.965          & 0.992          & 0.115            & 0.509          &   0.051                   \\
% Ren~\etal~\cite{Ren_2019_CVPR_Workshops}                                                        & 0.833          & 0.968          & 0.993          & 0.113            & 0.501          &   --             \\ 
% Zhang~\etal~\cite{Zhang_2019_CVPR}                                                      & 0.846          & 0.968          & 0.994          & 0.121            & 0.497          &   --             \\
SharpNet~\cite{Ramamonjisoa_2019_ICCV} & - & -                                                            & 0.836          & 0.966          & 0.993          & 0.139            & 0.502          &      {0.047}          \\
% Alhashim~\etal~\cite{Alhashim2018}                                                            & 0.846          & 0.974          & 0.994          & 0.123            & 0.465          &      0.053          \\
Hu~\etal~\cite{Hu2018RevisitingSI} & SENet-154  & 157                                                              & 0.866          & 0.975          & 0.993          & 0.115            & 0.530          &    0.050            \\ 
Chen~\etal~\cite{ijcai2019-98} & SENet & 210                                                      & 0.878          & 0.977          & 0.994          & 0.111            & 0.514          &  0.048              \\ 
Yin~\etal~\cite{Yin_2019_ICCV} & ResNeXt-101  & 114                                                             & 0.875          & 0.976          & 0.994          & {0.108}            & 0.416          & 0.048               \\ 
BTS~\cite{bts_lee2019big} & DenseNet-161   & 47                                                            & {0.885}          & 0.978          & 0.994          & 0.110            & {0.392}          & {0.047}          \\ 
%(Dont really want to cite them) DAV~\cite{dav_huynh2020guiding} & -  & 25                                                             & 0.882          & {0.980}          & {0.996} & {0.108}            & 0.412          & --              \\ 

AdaBins~\cite{Bhat2021} & EfficientNet-B5 & 78 & \underline{0.903} & \underline{0.984} & \underline{0.997} & \underline{0.103}     & \underline{0.364} & \underline{0.044} \\ 

\midrule
% The numbers here are from wandb-ECCV22-NYU, run_id : b8c0-
\textbf{LocalBins (Ours)}  & EfficientNet-B5 & 74 & \textbf{0.907} & \textbf{0.987} & \textbf{0.998} & \textbf{0.099}     & \textbf{0.357} & \textbf{0.042} \\ 
\bottomrule
\end{tabularx}
\end{adjustbox}
\caption{Comparison of performance on the NYU-Depth V2 dataset. The reported numbers are from the corresponding original papers. Best results are in bold, second best are underlined. }
\label{tab:results-nyu}
\end{table}

\subsection{Zero-shot performance}
To evaluate the generalization performance of our model, we use the model pretrained on the NYU-Depth V2 dataset and evaluate it on other datasets without fine-tuning.

\textbf{iBims-1 benchmark}~\cite{koch2019} (independent Benchmark images and matched scans version 1) is a high quality RGB-D dataset
acquired using a digital single-lens reflex (DSLR) camera and a high-precision laser scanner. Table.~\ref{tab:ibims} lists the performance on this benchmark, with our proposed model outperforming prior state-of-the-art methods. In addition, we show qualitative results in Fig.~\ref{fig:qualt_ibims}. AdaBins noticeably underestimates the depth range of the scenes with relative error growing with distance, whereas LocalBins more consistently predicts scale-accurate depths across varying depth ranges. This further emphasizes the generalization capability of our model.

\textbf{SUN-RGBD}~\cite{Song2015_sunrgbd} is an indoor dataset characterized by high scene diversity. We evaluate our model without fine-tuning on the official test set of 5050 images and report the results in Table.~\ref{tab:sunrgbd}.

\begin{table}[t]
\centering
\begin{tabular}{@{}lcccccc@{}}
\toprule
Method                    & \mdelta{1}            & \mdelta{2}            & \mdelta{3}            & REL$\downarrow$           & RMS$\downarrow$           & log10$\downarrow$         \\ \midrule
BTS                       & 0.54             & 0.86             & 0.95             & 0.23             & 0.93             & 0.11             \\
AdaBins                   & 0.55          & 0.87          & 0.96          & \textbf{0.21} & 0.90          & 0.11          \\ \midrule
\textbf{LocalBins (Ours)} & \textbf{0.56} & \textbf{0.88} & \textbf{0.97} & \textbf{0.21} & \textbf{0.88} & \textbf{0.10} \\ \bottomrule
\end{tabular}
\caption{Quantitative results on the iBims benchmark without fine-tuning.}
\label{tab:ibims}
\end{table}

\subsection{Analysis and Ablation Studies}
\label{subsec:analysis}
% Ablation table
\begin{table}[t]
\centering
\resizebox{0.7\linewidth}{!}{%
\begin{tabular}{@{}l|cccccccc@{}}
\toprule
  & Enc-Dec & LBM   & Naive & QR    & Lbins & Foveated & \textbf{REL} & \textbf{RMS} \\ \midrule
1 & \cmark    & \xmark & \xmark & \xmark & \xmark & \xmark    & 0.111        & 0.419        \\ 
2 & \cmark    & \cmark  & \xmark & \xmark & \xmark & \xmark    & 0.106        & 0.375        \\ \cmidrule(l){2-9} 
3 & \cmark    & \cmark  & $3\times3$  & \xmark & \cmark  & \xmark    & 0.108        & 0.381        \\
4 & \cmark    & \cmark  & $5\times5$  & \xmark & \cmark  & \xmark    & 0.107        & 0.375        \\
5 & \cmark    & \cmark  & $15\times15$  & \xmark & \cmark  & \xmark    & 0.108        & 0.377        \\ \cmidrule(l){2-9} 
8 & \cmark    & \cmark  & \xmark & \cmark  & \cmark  & \xmark    & 0.099        & 0.364        \\
9 & \cmark    & \cmark  & \xmark & \cmark  & \cmark  & \cmark     & 0.099        & 0.357        \\ \bottomrule
\end{tabular}
}
\caption{Ablation experiments showing the importance of various components in our proposed model. \textbf{Enc-Dec}: Base encoder-decoder model, \textbf{LBM}: LocalBins module, \textbf{Naive}: Naive training strategy discussed in Sec.~\ref{sec:training}, \textbf{QR}: Query-Response training, $\mathcal{L}_{bins}$: Chamfer loss supervision, \textbf{Foveated}: Foveated weighting in Chamfer loss. Data in the `Naive' column indicates the bbox size used on GT to compute density.}
\label{tab:ablation-main}
\end{table}

Here, we present the results of the extensive experiments we performed to analyse the properties and the importance of various components in our proposed model. 

\subsubsection{LocalBins module.}
We first evaluate the importance of our LocalBins module. We remove the LocalBins module from the network and evaluate our base encoder-decoder architecture. We use the pixel-wise loss ($\mathcal{L}_{pixel}$) to train the network. We also evaluate our proposed model (with LocalBins module) without the Chamfer loss to study the capacity of our design in absence of extra supervision. Results are reported in Table.~\ref{tab:ablation-main}.

\begin{table}[t]
\centering
\begin{tabular}{@{}c|cc|cl@{}}
\toprule
\multicolumn{1}{l}{}            & PSCI & \#px & \textbf{Coverage (\%)}~$\uparrow$ & \textbf{REL}~$\downarrow$    \\ \midrule
\multirow{2}{*}{Naive}          & 4096 & 4096           & 1.33     & 0.1075 \\
                                & 8192 & 8192           & 2.67     & 0.1071 \\ \midrule
\multirow{3}{*}{Query-Response} & 250  & 52,130         & 16.97    & 0.1043 \\
                                & 500  & 104,260        & 33.94    & 0.1002 \\
                                & 1000 & 208,520        & 67.88    & 0.0992  \\ \bottomrule
\end{tabular}
\caption{Quantitative demonstration of the efficiency of Query-Response Training. \textbf{PSCI}: Point Set Comparisons per Image. \textbf{\#px}: Total number of pixels covered for loss computation. \textbf{Coverage}: Percentage of \#px with respect to image resolution.}
\label{tab:coverage}
\end{table}

\begin{table}[t]
\centering
\begin{tabular}{@{}lllllll@{}}
\toprule
Method        & $\delta_1\uparrow$     & $\delta_2\uparrow$             & $\delta_3\uparrow$          & REL$\downarrow$         & RMS$\downarrow$  & $log_{10}\downarrow$\\ \midrule
Chen~\cite{ijcai2019-98}        & 0.757          & 0.943          & \underline{0.984}        & 0.166          & 0.494  &  0.071\\
Yin~\cite{Yin_2019_ICCV}        & 0.696          & 0.912          & 0.973          & 0.183          &  0.541  &  0.082\\
BTS~\cite{bts_lee2019big}            & 0.740          & 0.933          & 0.980          & 0.172          & 0.515  &  0.075 \\ 
AdaBins~\cite{Bhat2021}                       & \underline{0.771}  & \underline{0.944} & 0.983 & \underline{0.159} & \underline{0.476}  &  \underline{0.068}\\ \midrule
\textbf{Ours} & \textbf{0.777}  & \textbf{0.949} & \textbf{0.985} & \textbf{0.156} & \textbf{0.470}  &  \textbf{0.067}\\ \midrule
\bottomrule
\end{tabular}
\caption{Results of models trained on the NYU-Depth V2 dataset and tested on the SUN RGB-D dataset \cite{Song2015_sunrgbd} without fine-tuning.}
\label{tab:sunrgbd}
\end{table}

\subsubsection{Query-Response training and Foveated loss.}
We evaluate our proposed `Query-Response' training scheme against the naive implementation discussed in Sec.~\ref{sec:training}. As listed in Table.~\ref{tab:ablation-main}, Query-Response training gives a significant boost to the performance, improving the absolute relative error by \appx7\%. We believe the Query-Response training scheme is a general, powerful regularization technique that can be directly used in tasks beyond depth estimation. The foveated weighting scheme further improves the squared error based metrics.

%\begin{wraptable}{L}{5.5cm}
\begin{table}[t]
\centering
\begin{tabular}{@{}l|cc@{}}
\toprule
Splitter activation & \textbf{REL} & \textbf{RMS} \\ \midrule
Constant      & 0.117        & 0.454        \\
Sigmoid       & 0.100        & 0.361        \\
Linear norm         & 0.099        & 0.364        \\ \bottomrule
\end{tabular}
\caption{Types of splitters.}
\label{tab:splitters}
\end{table}
%\end{wraptable}

In order to demonstrate the power of Query-Response training, we compare it with the Naive scheme in terms of computation of Chamfer loss against the coverage (total number of pixel locations involved in loss computation - higher the better). We define `PSCI' as the total number of \textit{Point Set Comparisons performed per Image} and use it as an indicator of computational efficiency. Note that in practice, the computations are batched, and PSCI indicates the total number of `samples' in the batch contributed per image.
For the Naive scheme, PSCI is equal to the total number of subsampled locations. For the Query-Response training scheme, PSCI is equal to the total number of bounding box queries per image. We take random bounding boxes of sizes \{$3\times3$, $7\times7$, $15\times15$, $31\times31$, $63\times63$\} and compute their average total area. The results are given in Table.~\ref{tab:coverage}. Our proposed training scheme performs \appx7.6\% better with $8\times$ fewer point set comparisons compared to the naive scheme.

\subsubsection{Splitter activation function.} We evaluate the three types of splitter activation functions as discussed in Sec.~\ref{subsec:arch}. The results are given in Table.~\ref{tab:splitters}.

\subsubsection{Effect of $N_{seed}$.}
\begin{figure}[t]
    \centering
    \includegraphics[width=0.6\linewidth]{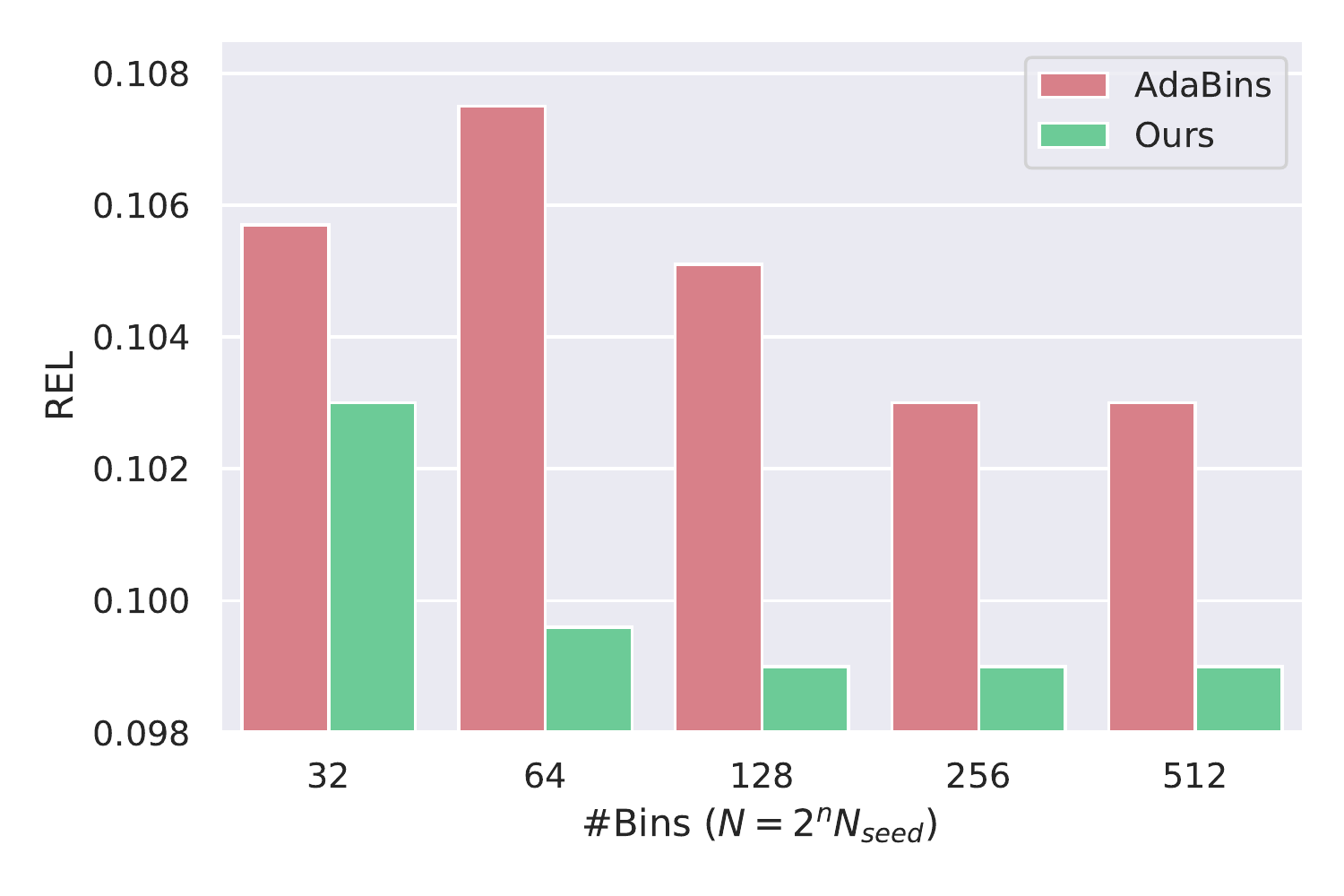}
    \caption{REL vs \#Bins.}
    \label{fig:num_bins}
\end{figure}

We analyse the effect of varying the $N_{seed}$ and hence the total number of bins, and compare with AdaBins. The results are plotted in Fig.~\ref{fig:num_bins}. We find that our model is more robust to the total number of bins used and generally has better performance.

% To be moved to supplementary
% \subsubsection{Investigating `region of influence'.}
% To provide some idea of the inner workings of LocalBins, we provide a visualization of XXX in Table~\ref{}.

% \begin{figure}[h]
%     \centering
%     \includegraphics[width=0.6\linewidth]{example-image-a}
%     \caption{Visualization of area of local neighborhood that influences the bin prediction}
%     \label{fig:my_label}
% \end{figure}
\section{Conclusions}
We introduced a new network architecture for depth estimation from a single image. We build on an encoder-decoder architecture and evolve the current state-of-the-art model AdaBins in two aspects. First, we add three building blocks to estimate local depth distributions in the neighborhood of a pixel. These three building blocks, bin embedding layers, seed bin width estimator, and bin splitters are tightly integrated with the decoder in a multi-scale fashion. Second, we propose a query - response acceleration strategy for training, since a naive implementation of the idea would be highly time and memory consuming.
In future work, we would like to adapt the LocalBins concept to other dense regression algorithms, such as image segmentation or inpainting.

% The paper ends with a conclusion. 

% This is the last page of the manuscript.
% Now we have reached the maximum size of the ECCV 2022 submission (excluding references).
% References should start immediately after the main text, but can continue on p.15 if needed.

\clearpage
\bibliographystyle{splncs04}
\bibliography{egbib}

\begin{thebibliography}{10}
\providecommand{\url}[1]{\texttt{#1}}
\providecommand{\urlprefix}{URL }
\providecommand{\doi}[1]{https://doi.org/#1}

\bibitem{AkadaBAW22}
Akada, H., Bhat, S.F., Alhashim, I., Wonka, P.: Self-supervised learning of
  domain invariant features for depth estimation. In: {IEEE/CVF} Winter
  Conference on Applications of Computer Vision, {WACV} 2022, Waikoloa, HI,
  USA, January 3-8, 2022. pp. 997--1007. {IEEE} (2022).
  \doi{10.1109/WACV51458.2022.00107},
  \url{https://doi.org/10.1109/WACV51458.2022.00107}

\bibitem{Alhashim2018}
Alhashim, I., Wonka, P.: High quality monocular depth estimation via transfer
  learning. CoRR  \textbf{abs/1812.11941} (2018),
  \url{http://arxiv.org/abs/1812.11941}

\bibitem{atapour2018real}
Atapour-Abarghouei, A., Breckon, T.P.: Real-time monocular depth estimation
  using synthetic data with domain adaptation via image style transfer. In:
  Proceedings of the IEEE Conference on Computer Vision and Pattern
  Recognition. pp. 2800--2810 (2018)

\bibitem{Bhat2021}
Bhat, S.F., Alhashim, I., Wonka, P.: Adabins: Depth estimation using adaptive
  bins. In: 2021 IEEE/CVF Conference on Computer Vision and Pattern Recognition
  (CVPR). pp. 4008--4017. IEEE Computer Society, Los Alamitos, CA, USA (jun
  2021). \doi{10.1109/CVPR46437.2021.00400},
  \url{https://doi.ieeecomputersociety.org/10.1109/CVPR46437.2021.00400}

\bibitem{casser2019unsupervised}
Casser, V., Pirk, S., Mahjourian, R., Angelova, A.: Unsupervised monocular
  depth and ego-motion learning with structure and semantics. In: CVPR Workshop
  on Visual Odometry and Computer Vision Applications Based on Location Cues
  (VOCVALC) (2019)

\bibitem{ijcai2019-98}
Chen, X., Chen, X., Zha, Z.J.: Structure-aware residual pyramid network for
  monocular depth estimation. In: Proceedings of the Twenty-Eighth
  International Joint Conference on Artificial Intelligence, {IJCAI-19}. pp.
  694--700. International Joint Conferences on Artificial Intelligence
  Organization (7 2019). \doi{10.24963/ijcai.2019/98},
  \url{https://doi.org/10.24963/ijcai.2019/98}

\bibitem{CrDoCo2019}
Chen, Y.C., Lin, Y.Y., Yang, M.H., Huang, J.B.: Crdoco: Pixel-level domain
  transfer with cross-domain consistency. In: IEEE Conference on Computer
  Vision and Pattern Recognition (CVPR) (2019)

\bibitem{Eigen2014}
Eigen, D., Puhrsch, C., Fergus, R.: Depth map prediction from a single image
  using a multi-scale deep network. In: NIPS (2014)

\bibitem{Fu2018DeepOR}
Fu, H., Gong, M., Wang, C., Batmanghelich, N., Tao, D.: Deep ordinal regression
  network for monocular depth estimation. 2018 IEEE/CVF Conference on Computer
  Vision and Pattern Recognition pp. 2002--2011 (2018)

\bibitem{Godard2017}
Godard, C., Aodha, O.M., Brostow, G.J.: Unsupervised monocular depth estimation
  with left-right consistency. 2017 IEEE Conference on Computer Vision and
  Pattern Recognition (CVPR) pp. 6602--6611 (2017)

\bibitem{Godard2018DiggingIS}
Godard, C., Aodha, O.M., Brostow, G.J.: Digging into self-supervised monocular
  depth estimation. CoRR  \textbf{abs/1806.01260} (2018)

\bibitem{Gordon_2019_ICCV}
Gordon, A., Li, H., Jonschkowski, R., Angelova, A.: Depth from videos in the
  wild: Unsupervised monocular depth learning from unknown cameras. In:
  Proceedings of the IEEE/CVF International Conference on Computer Vision
  (ICCV) (October 2019)

\bibitem{Hao2018DetailPD}
Hao, Z., Li, Y., You, S., Lu, F.: Detail preserving depth estimation from a
  single image using attention guided networks. 2018 International Conference
  on 3D Vision (3DV) pp. 304--313 (2018)

\bibitem{He2017_MaskRCNN}
He, K., Gkioxari, G., Dollár, P., Girshick, R.: Mask r-cnn. In: 2017 IEEE
  International Conference on Computer Vision (ICCV). pp. 2980--2988 (2017).
  \doi{10.1109/ICCV.2017.322}

\bibitem{Hu2018RevisitingSI}
Hu, J., Ozay, M., Zhang, Y., Okatani, T.: Revisiting single image depth
  estimation: Toward higher resolution maps with accurate object boundaries.
  2019 IEEE Winter Conference on Applications of Computer Vision (WACV) pp.
  1043--1051 (2018)

\bibitem{dav_huynh2020guiding}
Huynh, L., Nguyen-Ha, P., Matas, J., Rahtu, E., Heikkila, J.: Guiding monocular
  depth estimation using depth-attention volume. arXiv preprint
  arXiv:2004.02760  (2020)

\bibitem{kim2022global}
Kim, D., Ga, W., Ahn, P., Joo, D., Chun, S., Kim, J.: Global-local path
  networks for monocular depth estimation with vertical cutdepth. arXiv
  preprint arXiv:2201.07436  (2022)

\bibitem{koch2019}
Koch, T., Liebel, L., Fraundorfer, F., K{\"o}rner, M.: Evaluation of cnn-based
  single-image depth estimation methods. In: Proceedings ECCV 2018 Workshops
  (2019)

\bibitem{Laina2016}
Laina, I., Rupprecht, C., Belagiannis, V., Tombari, F., Navab, N.: Deeper depth
  prediction with fully convolutional residual networks. 2016 Fourth
  International Conference on 3D Vision (3DV) pp. 239--248 (2016)

\bibitem{bts_lee2019big}
Lee, J.H., Han, M.K., Ko, D.W., Suh, I.H.: From big to small: Multi-scale local
  planar guidance for monocular depth estimation. arXiv preprint
  arXiv:1907.10326  (2019)

\bibitem{Lee2011}
Lee, W., Park, N., Woo, W.: Depth-assisted real-time 3d object detection for
  augmented reality. ICAT’11  \textbf{2},  126--132 (2011)

\bibitem{li2020unsupervised}
Li, H., Gordon, A., Zhao, H., Casser, V., Angelova, A.: Unsupervised monocular
  depth learning in dynamic scenes. arXiv preprint arXiv:2010.16404  (2020)

\bibitem{MDLi18}
Li, Z., Snavely, N.: Megadepth: Learning single-view depth prediction from
  internet photos. In: Computer Vision and Pattern Recognition (CVPR) (2018)

\bibitem{LoshchilovH19}
Loshchilov, I., Hutter, F.: Decoupled weight decay regularization. In: 7th
  International Conference on Learning Representations, {ICLR} 2019, New
  Orleans, LA, USA, May 6-9, 2019. OpenReview.net (2019),
  \url{https://openreview.net/forum?id=Bkg6RiCqY7}

\bibitem{NEURIPS2019_bdbca288}
Paszke, A., Gross, S., Massa, F., Lerer, A., Bradbury, J., Chanan, G., Killeen,
  T., Lin, Z., Gimelshein, N., Antiga, L., Desmaison, A., Kopf, A., Yang, E.,
  DeVito, Z., Raison, M., Tejani, A., Chilamkurthy, S., Steiner, B., Fang, L.,
  Bai, J., Chintala, S.: Pytorch: An imperative style, high-performance deep
  learning library. In: Wallach, H., Larochelle, H., Beygelzimer, A.,
  d\textquotesingle Alch\'{e}-Buc, F., Fox, E., Garnett, R. (eds.) Advances in
  Neural Information Processing Systems. vol.~32, pp. 8026--8037. Curran
  Associates, Inc. (2019),
  \url{https://proceedings.neurips.cc/paper/2019/file/bdbca288fee7f92f2bfa9f7012727740-Paper.pdf}

\bibitem{Ramamonjisoa_2019_ICCV}
Ramamonjisoa, M., Lepetit, V.: Sharpnet: Fast and accurate recovery of
  occluding contours in monocular depth estimation. In: Proceedings of the
  IEEE/CVF International Conference on Computer Vision (ICCV) Workshops (Oct
  2019)

\bibitem{Ranftl_2021_ICCV}
Ranftl, R., Bochkovskiy, A., Koltun, V.: Vision transformers for dense
  prediction. In: Proceedings of the IEEE/CVF International Conference on
  Computer Vision (ICCV). pp. 12179--12188 (October 2021)

\bibitem{Ranftl2020MiDaS}
Ranftl, R., Lasinger, K., Hafner, D., Schindler, K., Koltun, V.: Towards robust
  monocular depth estimation: Mixing datasets for zero-shot cross-dataset
  transfer. IEEE Transactions on Pattern Analysis and Machine Intelligence
  (TPAMI)  (2020)

\bibitem{Silberman2012}
Silberman, N., Hoiem, D., Kohli, P., Fergus, R.: Indoor segmentation and
  support inference from rgbd images. In: Computer Vision -- ECCV 2012. pp.
  746--760. Springer Berlin Heidelberg, Berlin, Heidelberg (2012)

\bibitem{Song2015_sunrgbd}
{Song}, S., {Lichtenberg}, S.P., {Xiao}, J.: Sun rgb-d: A rgb-d scene
  understanding benchmark suite. In: 2015 IEEE Conference on Computer Vision
  and Pattern Recognition (CVPR). pp. 567--576 (2015).
  \doi{10.1109/CVPR.2015.7298655}

\bibitem{Tonioni2019}
Tonioni, A., Poggi, M., Mattoccia, S., di~Stefano, L.: Unsupervised domain
  adaptation for depth prediction from images. CoRR  \textbf{abs/1909.03943}
  (2019), \url{http://arxiv.org/abs/1909.03943}

\bibitem{Watson_2021_CVPR}
Watson, J., Mac~Aodha, O., Prisacariu, V., Brostow, G., Firman, M.: The
  temporal opportunist: Self-supervised multi-frame monocular depth. In:
  Proceedings of the IEEE/CVF Conference on Computer Vision and Pattern
  Recognition (CVPR). pp. 1164--1174 (June 2021)

\bibitem{Xie2016Deep3DFA}
Xie, J., Girshick, R.B., Farhadi, A.: Deep3d: Fully automatic 2d-to-3d video
  conversion with deep convolutional neural networks. In: ECCV (2016)

\bibitem{Xu2017}
Xu, D., Ricci, E., Ouyang, W., Wang, X., Sebe, N.: Multi-scale continuous crfs
  as sequential deep networks for monocular depth estimation. In: Proceedings
  of the IEEE Conference on Computer Vision and Pattern Recognition. pp.
  5354--5362 (2017)

\bibitem{Xu2018StructuredAG}
Xu, D., Wang, W., Tang, H., Liu, H.W., Sebe, N., Ricci, E.: Structured
  attention guided convolutional neural fields for monocular depth estimation.
  2018 IEEE/CVF Conference on Computer Vision and Pattern Recognition pp.
  3917--3925 (2018)

\bibitem{Yin_2019_ICCV}
Yin, W., Liu, Y., Shen, C., Yan, Y.: Enforcing geometric constraints of virtual
  normal for depth prediction. In: Proceedings of the IEEE/CVF International
  Conference on Computer Vision (ICCV) (October 2019)

\bibitem{Yuan2022_cvpr_crf}
{Yuan}, W., {Gu}, X., {Dai}, Z., {Zhu}, S., {Tan}, P.: {NeW CRFs: Neural Window
  Fully-connected CRFs for Monocular Depth Estimation}. arXiv e-prints
  arXiv:2203.01502 (Mar 2022)

\bibitem{zhao2019geometry}
Zhao, S., Fu, H., Gong, M., Tao, D.: Geometry-aware symmetric domain adaptation
  for monocular depth estimation. In: Proceedings of the IEEE/CVF Conference on
  Computer Vision and Pattern Recognition. pp. 9788--9798 (2019)

\bibitem{Zhao_2020_CVPR}
Zhao, Y., Kong, S., Shin, D., Fowlkes, C.: Domain decluttering: Simplifying
  images to mitigate synthetic-real domain shift and improve depth estimation.
  In: Proceedings of the IEEE/CVF Conference on Computer Vision and Pattern
  Recognition (CVPR) (June 2020)

\bibitem{Zhou2021Diffnet}
Zhou, H., Greenwood, D., Taylor, S.: Self-supervised monocular depth estimation
  with internal feature fusion. In: British Machine Vision Conference (BMVC)
  (2021)

\bibitem{Zhou2017}
Zhou, T., Brown, M.R., Snavely, N., Lowe, D.G.: Unsupervised learning of depth
  and ego-motion from video. 2017 IEEE Conference on Computer Vision and
  Pattern Recognition (CVPR) pp. 6612--6619 (2017)

\end{thebibliography}

%%%%%%%%%%%%%%%%%% Supplementary material %%%%%%%%%%%%%%%%%%%%%%%%%%%%

\clearpage
\appendix
\section{Appendix}
\subsection{Visualizing bin predictions}
\begin{figure}
    \centering
    \includegraphics[width=\linewidth]{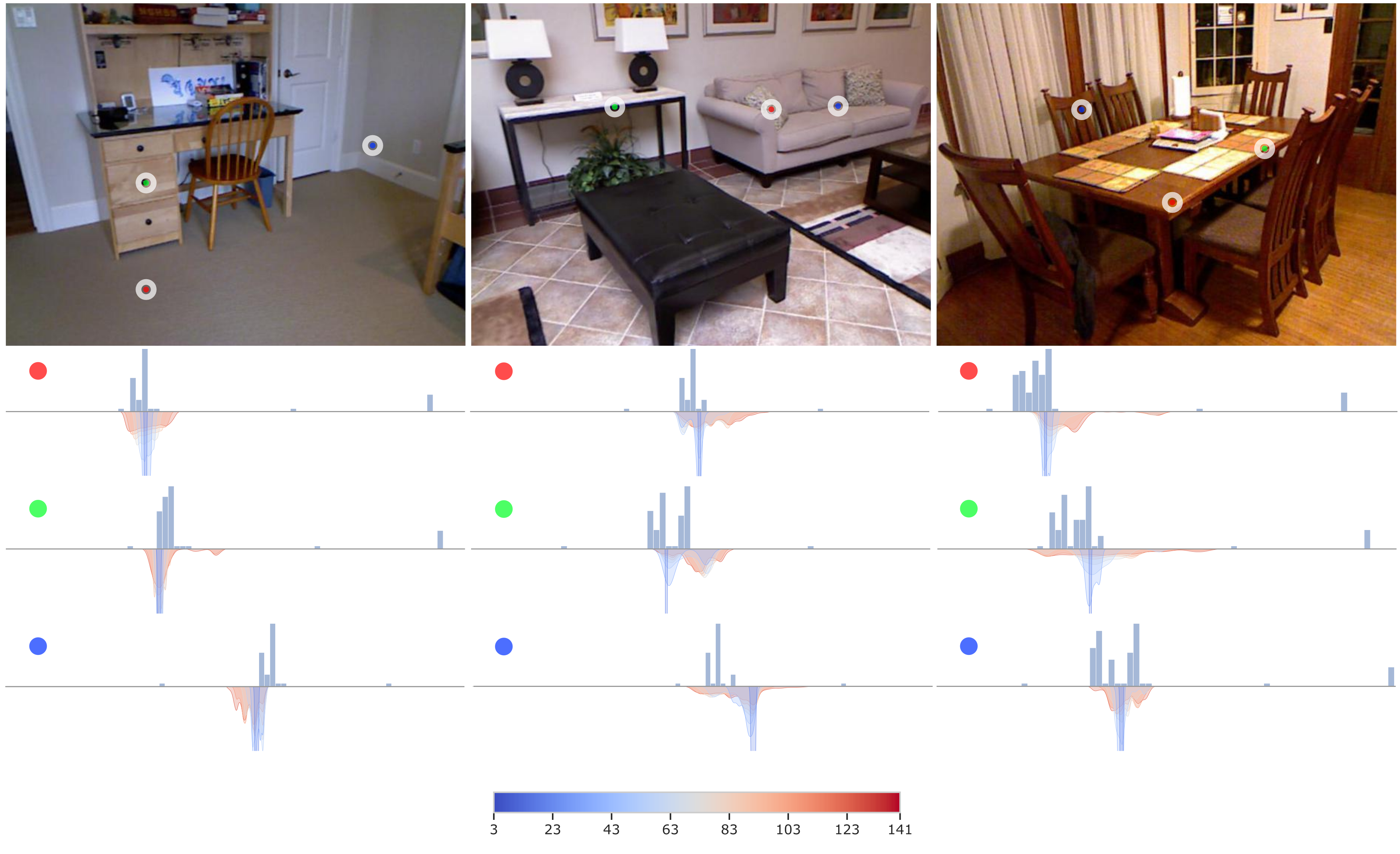}
    \caption{Visualization of bin predictions, showing input RGB (first row) and density plots of local depth distributions at selected locations across the depth range interval. Density plots consisting of density of bin centers predicted by LocalBins module at the selected pixel location (\textbf{top}) and density of ground truth depth values (\textbf{bottom}) for various window sizes (indicated by the colorbar).}
    \label{fig:qualt_bins}
\end{figure}

Fig.~\ref{fig:qualt_bins} shows a qualitative comparison of the bin predictions of the LocalBins module against the ground truth depth distribution in the local neighborhoods of various sizes. 

\begin{figure}[t]
    \centering
    \includegraphics[width=\linewidth]{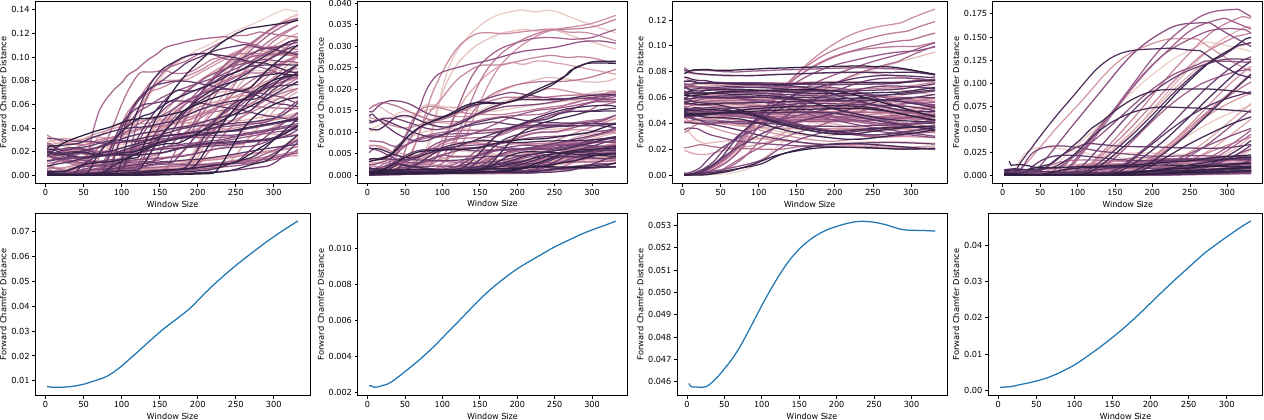}
    \caption{How \textit{``local"} are the bin predictions of the LocalBins module? Plots show the average absolute difference between the ground truth depth values and their nearest bin centers predicted by the LocalBins module. Top row shows the average differences plotted for 100 random locations for four randomly selected input images (columns). GT depth values are taken from the `concentric' bounding boxes and compared against the bins predicted at the center. Bottom row shows the mean across the 100 locations for each image.}
    \label{fig:plot_dists}
\end{figure}

Our main goal in this work was to have the LocalBins module predict the local depth distribution at each pixel. However, in Query-Response training, windows of various sizes are generated, and regularization is imposed on the bin predictions of the entire window together rather than for each individual pixel location separately. While being vital (refer to `Query-Response training' section in the main paper), this renders the `size' of the `local neighborhood' rather indeterministic. The model may either end up using a fixed size and always predict the bins at each pixel that reflect the density of depth values within a fixed-sized neighborhood, or the model may as well choose to cover different sized neighborhoods depending upon the context. Visualizations (See Fig.~\ref{fig:plot_dists}) indicate the latter case. 

\subsection{Different Backbones}

% \usepackage{booktabs}
% \usepackage{graphicx}
\begin{table}[]
\centering
\resizebox{\textwidth}{!}{%
\begin{tabular}{@{}lc|llllll@{}}
\toprule
\multicolumn{1}{c}{Backbone} &
  \multicolumn{1}{c|}{\#params(M)} &
  \multicolumn{1}{c}{d1} &
  \multicolumn{1}{c}{d2} &
  \multicolumn{1}{c}{d3} &
  \multicolumn{1}{c}{REL} &
  RMS &
  log10 \\ \midrule
MobileNetV2-100  & 20.26 & 0.812 & 0.963 & 0.992 & 0.141 & 0.480 & 0.060 \\
EfficientNetV2-S & 38.71 & 0.894 & 0.981 & 0.995 & 0.106 & 0.376 & 0.045 \\
EfficientNetV2-M & 71.53 & 0.898 & 0.983 & 0.996 & 0.102 & 0.365 & 0.044 \\
EfficientNetV2-L & 136.3 & 0.910 & 0.986 & 0.997 & 0.098 & 0.351 & 0.042 \\ \bottomrule
\end{tabular}%
}
\caption{Performance of LocalBins with various backbone encoders}
\label{tab:backbones}
\end{table}

We switch the EfficientNet-b5 encoder in our default model with various other backbones and list the performance in Table.~\ref{tab:backbones}

% The paper ends with a conclusion. 

% This is the last page of the manuscript.
% Now we have reached the maximum size of the ECCV 2022 submission (excluding references).
% References should start immediately after the main text, but can continue on p.15 if needed.

\clearpage

\end{document}

% \linenumbers
\pagestyle{headings}
\mainmatter
\def\ECCVSubNumber{2871}  % Insert your submission number here

% Proposed title
\title{Appendix - LocalBins: Improving Depth Estimation by Learning Local Distributions} % Replace with your title

%\begin{comment}
\titlerunning{ECCV-22 submission ID \ECCVSubNumber} 
\authorrunning{ECCV-22 submission ID \ECCVSubNumber} 
\author{Anonymous ECCV submission}
\institute{Paper ID \ECCVSubNumber}

%\end{comment}
%******************

\begin{comment}
\titlerunning{Abbreviated paper title}
\author{First Author\inst{1}\orcidID{0000-1111-2222-3333} \and
Second Author\inst{2,3}\orcidID{1111-2222-3333-4444} \and
Third Author\inst{3}\orcidID{2222--3333-4444-5555}}
%
\authorrunning{F. Author et al.}
\institute{Princeton University, Princeton NJ 08544, USA \and
Springer Heidelberg, Tiergartenstr. 17, 69121 Heidelberg, Germany
\email{lncs@springer.com}\\
\url{http://www.springer.com/gp/computer-science/lncs} \and
ABC Institute, Rupert-Karls-University Heidelberg, Heidelberg, Germany\\
\email{\{abc,lncs\}@uni-heidelberg.de}}
\end{comment}
%******************
\maketitle

\section{Visualizing bin predictions}
\begin{figure}
    \centering
    \includegraphics[width=\linewidth]{root/figures/binviz_v3.png}
    \caption{Visualization of bin predictions, showing input RGB (first row) and density plots of local depth distributions at selected locations across the depth range interval. Density plots consisting of density of bin centers predicted by LocalBins module at the selected pixel location (\textbf{top}) and density of ground truth depth values (\textbf{bottom}) for various window sizes (indicated by the colorbar).}
    \label{fig:qualt_bins}
\end{figure}

Fig.~\ref{fig:qualt_bins} shows a qualitative comparison of the bin predictions of the LocalBins module against the ground truth depth distribution in the local neighborhoods of various sizes. 

Our main goal in this work was to have the LocalBins module predict the local depth distribution at each pixel. However, in Query-Response training, windows of various sizes are generated, and regularization is imposed on the bin predictions of the entire window together rather than for each individual pixel location separately. While being vital (refer to `Query-Response training' section in the main paper), this renders the `size' of the `local neighborhood' rather indeterministic. The model may either end up using a fixed size and always predict the bins at each pixel that reflect the density of depth values within a fixed-sized neighborhood, or the model may as well choose to cover different sized neighborhoods depending upon the context. Visualizations (See Fig.~\ref{fig:plot_dists}) indicate the latter case. 

\begin{figure}[t]
    \centering
    \includegraphics[width=\linewidth]{root/figures/cdist_plots_v2.pdf}
    \caption{How \textit{``local"} are the bin predictions of the LocalBins module? Plots show the average absolute difference between the ground truth depth values and their nearest bin centers predicted by the LocalBins module. Top row shows the average differences plotted for 100 random locations for four randomly selected input images (columns). GT depth values are taken from the `concentric' bounding boxes and compared against the bins predicted at the center. Bottom row shows the mean across the 100 locations for each image.}
    \label{fig:plot_dists}
\end{figure}

\section{Different Backbones}

% \usepackage{booktabs}
% \usepackage{graphicx}
\begin{table}[]
\centering
\resizebox{\textwidth}{!}{%
\begin{tabular}{@{}lc|llllll@{}}
\toprule
\multicolumn{1}{c}{Backbone} &
  \multicolumn{1}{c|}{\#params(M)} &
  \multicolumn{1}{c}{d1} &
  \multicolumn{1}{c}{d2} &
  \multicolumn{1}{c}{d3} &
  \multicolumn{1}{c}{REL} &
  RMS &
  log10 \\ \midrule
MobileNetV2-100  & 20.26 & 0.812 & 0.963 & 0.992 & 0.141 & 0.480 & 0.060 \\
EfficientNetV2-S & 38.71 & 0.894 & 0.981 & 0.995 & 0.106 & 0.376 & 0.045 \\
EfficientNetV2-M & 71.53 & 0.898 & 0.983 & 0.996 & 0.102 & 0.365 & 0.044 \\
EfficientNetV2-L & 136.3 & 0.910 & 0.986 & 0.997 & 0.098 & 0.351 & 0.042 \\ \bottomrule
\end{tabular}%
}
\caption{Performance of LocalBins with various backbone encoders}
\label{tab:backbones}
\end{table}

We switch the EfficientNet-b5 encoder in our default model with various other backbones and list the performance in Table.~\ref{tab:backbones}

% The paper ends with a conclusion. 

% This is the last page of the manuscript.
% Now we have reached the maximum size of the ECCV 2022 submission (excluding references).
% References should start immediately after the main text, but can continue on p.15 if needed.

\clearpage
% % References will then be sorted and formatted in the correct style.
% \bibliography{egbib}